\documentclass[11pt,a4paper]{article}
\usepackage[hyperref]{acl2020}
\usepackage{times}
\usepackage{latexsym}

\usepackage{times}
\usepackage{url}

\usepackage{latexsym}
\usepackage{amsmath}
\usepackage{multirow}
\usepackage[multiple]{footmisc}
\usepackage{array, booktabs, makecell}
\usepackage{graphicx}
\usepackage{colortbl}
\usepackage{xcolor}
\usepackage{microtype}

\newcommand{\sm}[1]{\textcolor{red}{\bf\small [#1 --SM]}}

\newcommand{\Sref}[1]{\S\ref{#1}}
\def\V#1{\mathbf{#1}}
\def\C#1{\mathcal{#1}}

\aclfinalcopy 
\def\aclpaperid{2505} 

\newcommand\datasetsize{1.39 million }
\title{Politeness Transfer: A Tag and Generate Approach}

\author{Aman Madaan~\thanks{\hspace{0.5em} authors contributed equally to this work.}\hspace{0.5em}, Amrith Setlur~\footnotemark[1]\hspace{0.5em}, Tanmay Parekh~\footnotemark[1]\hspace{0.5em}, Barnabas Poczos, Graham Neubig,\\
\textbf{Yiming Yang, Ruslan Salakhutdinov, Alan W Black, Shrimai Prabhumoye}  \\
  School of Computer Science \\
  Carnegie Mellon University \\
  Pittsburgh, PA, USA \\
  \texttt{\{amadaan, asetlur, tparekh\}@cs.cmu.edu} \\}

\date{}

\begin{document}
\maketitle
\begin{abstract}
This paper introduces a new task of politeness transfer which involves converting non-polite sentences to polite sentences while preserving the meaning.
We also provide a dataset of more than \datasetsize  instances automatically labeled for politeness to encourage benchmark evaluations on this new task. 
We design a \emph{tag} and \emph{generate} pipeline that identifies stylistic attributes and subsequently generates a sentence in the target style while preserving most of the source content.
For politeness as well as five other transfer tasks, our model outperforms the state-of-the-art methods on automatic metrics for content preservation, with a  comparable or better performance on style transfer accuracy.
Additionally, our model surpasses existing methods on human evaluations for grammaticality, meaning preservation and transfer accuracy across all the six style transfer tasks.
The data and code is located at \url{https://github.com/tag-and-generate/}
\end{abstract}

\section{Introduction}
\label{sec:introduction}





Politeness plays a crucial role in social interaction, and is closely tied with power dynamics, social distance between the participants of a conversation, and gender \cite{brown1987politeness, danescu-niculescu-mizil-etal-2013-computational}.
It is also imperative to use the appropriate level of politeness for smooth communication in conversations \cite{coppock2005politeness}, organizational settings like emails \cite{peterson-etal-2011-email}, memos, official documents, and many other settings.
Notably, politeness has also been identified as an interpersonal style which can be decoupled from content \cite{kang2019xslue}.
Motivated by its central importance, in this paper we study the task of converting non-polite sentences to polite sentences while preserving the meaning.

Prior work on text style transfer \cite{shen2017style, li-etal-2018-delete, prabhumoye-etal-2018-style, rao2018dear, xu-etal-2012-paraphrasing, jhamtani-etal-2017-shakespearizing} has not focused on politeness as a style transfer task, and we argue that defining it is cumbersome. 
While native speakers of a language and cohabitants of a region have a good working understanding of the phenomenon of politeness for everyday conversation, pinning it down as a definition is non-trivial \cite{meier1995defining}. 
There are primarily two reasons for this complexity. First, as noted by \cite{brown1987politeness}, the phenomenon of politeness is rich and multifaceted.
Second, politeness of a sentence depends on the culture, language, and social structure of both the speaker and the addressed person. For instance, while using ``please'' in requests made to the closest friends is common amongst the native speakers of North American English, such an act would be considered awkward, if not rude, in the Arab culture \cite{kadar2011politeness}.  

We circumscribe the scope of politeness for the purpose of this study as follows:
First, we adopt the data driven definition of politeness proposed by  \cite{danescu-niculescu-mizil-etal-2013-computational}.
Second, we base our experiments on a dataset derived from the Enron corpus \cite{klimt2004introducing} which consists of email exchanges in an American corporation. Thus, we restrict our attention to the notion of politeness as widely accepted by the speakers of North American English in a formal setting.

Even after framing politeness transfer as a task, there are additional challenges involved that differentiate politeness from other styles. Consider a common directive in formal communication, ``send me the data''. While the sentence is not impolite, a rephrasing ``could you please send me the data'' would largely be accepted as a more polite way of phrasing the same statement \citep{danescu-niculescu-mizil-etal-2013-computational}.
This example brings out a distinct characteristic of politeness. 
It is easy to pinpoint the signals for \emph{politeness}. 
However, cues that signal the \emph{absence} of politeness, like direct questions, statements and factuality \cite{danescu-niculescu-mizil-etal-2013-computational}, do not explicitly appear in a sentence, and are thus hard to objectify.
Further, the other extreme of politeness, impolite sentences, are typically riddled with curse words and insulting phrases. While interesting, such cases can typically be neutralized using lexicons.
For our study, we focus on the task of transferring the non-polite sentences to polite sentences, where we simply define non-politeness to be the absence of both politeness and impoliteness.
Note that this is in stark contrast with the standard style transfer tasks, which involve transferring a sentence from a well-defined style polarity to the other (like positive to negative sentiment).

We propose a \emph{tag} and \emph{generate} pipeline to overcome these challenges.
The \textit{tagger} identifies the words or phrases which belong to the original style and replaces them with a tag token.
If the sentence has no style attributes, as in the case for politeness transfer, the tagger adds the tag token in positions where phrases in the target style can be inserted.
The \textit{generator} takes as input the output of the tagger and generates a sentence in the target style. 
Additionally, unlike previous systems, the outputs of the intermediate steps in our system are fully realized, making the whole pipeline interpretable. Finally, if the input sentence is already in the target style, our model won't add any stylistic markers and thus would allow the input to flow as is.

We evaluate our model on politeness transfer as well as 5 additional tasks described in prior work \cite{shen2017style, prabhumoye-etal-2018-style, li-etal-2018-delete} on content preservation, fluency and style transfer accuracy.
Both automatic and human evaluations show that our model beats the state-of-the-art methods in content preservation, while either matching or improving the transfer accuracy across six different style transfer tasks(\Sref{sec:expt_results}).
The results show that our technique is effective across a broad spectrum of style transfer tasks.
Our methodology is inspired by \citet{li-etal-2018-delete} and improves upon several of its limitations as described in (\Sref{sec:related_work}).


Our main contribution is the design of politeness transfer task. 
To this end, we provide a large dataset of nearly \datasetsize sentences labeled for politeness (\url{https://github.com/tag-and-generate/politeness-dataset}).
Additionally, we hand curate a test set of 800 samples (from Enron emails) which are annotated as requests. To the best of our knowledge, we are the first to undertake politeness as a style transfer task. In the process, we highlight an important class of problems wherein the transfer involves going from a neutral style to the target style.
Finally, we design a ``tag and generate'' pipeline that is particularly well suited for tasks like politeness, while being general enough to match or beat the performance of the existing systems on popular style transfer tasks.

\section{Related Work}
\label{sec:related_work}
Politeness and its close relation with power dynamics and social interactions has been well documented \cite{brown1987politeness}.
Recent work \cite{danescu-niculescu-mizil-etal-2013-computational} in computational linguistics has provided a corpus of \textit{requests} annotated for politeness curated from Wikipedia and StackExchange.
\citeauthor{niu-bansal-2018-polite} \citeyearpar{niu-bansal-2018-polite} uses this corpus to generate polite dialogues.
Their work focuses on contextual dialogue response generation as opposed to content preserving style transfer, while the latter is the central theme of our work.
Prior work on Enron corpus \cite{yeh2006email} has been mostly from a socio-linguistic perspective to observe social power dynamics \cite{Bramsen:2011:ESP:2002472.2002570, McCallum:2007:TRD:1622637.1622644}, formality \cite{peterson-etal-2011-email} and politeness \cite{prabhakaran-etal-2014-gender}.
We build upon this body of work by using this corpus as a source for the style transfer task.

Prior work on style transfer has largely focused on tasks of sentiment modification \cite{pmlr-v70-hu17e, shen2017style, li-etal-2018-delete}, caption transfer \cite{li-etal-2018-delete}, persona transfer \cite{chandu2019my, zhang-etal-2018-personalizing}, gender and political slant transfer \cite{reddy2016obfuscating, prabhumoye-etal-2018-style}, and formality transfer \cite{rao2018dear, xu2019formality}.
Note that formality and politeness are loosely connected but independent styles \cite{kang2019xslue}. We focus our efforts on carving out a task for politeness transfer and creating a dataset for such a task. 


Current style transfer techniques \cite{shen2017style, pmlr-v70-hu17e, fu2018style, yang2018unsupervised, john-etal-2019-disentangled} try to disentangle source style from content and then combine the content with the target style to generate the sentence in the target style.
Compared to prior work, ``Delete, Retrieve and Generate'' \cite{li-etal-2018-delete} (referred to as \textsc{drg} henceforth) and its extension \cite{sudhakar-etal-2019-transforming} are effective methods to generate outputs in the target style while having a relatively high rate of source content preservation.
However, \textsc{drg} has several limitations: (1) the delete module often marks content words as stylistic markers and deletes them,
(2) the retrieve step relies on the presence of similar content in both the source and target styles,
(3) the retrieve step is time consuming for large datasets, 
(4) the pipeline makes the assumption that style can be transferred by deleting stylistic markers and replacing them with target style phrases,
(5) the method relies on a fixed corpus of style attribute markers, and is thus limited in its ability to generalize to unseen data during test time.
Our methodology differs from these works as it does not require the retrieve stage and makes no assumptions on the existence of similar content phrases in both the styles.
This also makes our pipeline faster in addition to being robust to noise.

\citet{ijcai2019-732} treats style transfer as a conditional language modelling task.
It focuses only on sentiment modification, treating it as a cloze form task of filling in the appropriate words in the target sentiment. 
In contrast, we are capable of generating the entire sentence in the target style.
Further, our work is more generalizable and we show results on five other style transfer tasks.

\section{Tasks and Datasets}


\subsection{Politeness Transfer Task}

For the politeness transfer task, we focus on sentences in which the speaker communicates a requirement that the listener needs to fulfill.
Common examples include imperatives ``\emph{Let's stay in touch}'' and questions that express a proposal ``\emph{Can you call me when you get back?}''.
Following \citeauthor{Jurafsky-etal:1997} \citeyearpar{Jurafsky-etal:1997}, we use the umbrella term ``action-directives'' for such sentences. The goal of this task is to convert action-directives to polite requests. 
While there can be more than one way of making a sentence polite, for the above examples, adding gratitude (``\textit{Thanks} and let's stay in touch'') or counterfactuals (``\emph{Could} you please call me when you get back?'') would make them polite \cite{danescu-niculescu-mizil-etal-2013-computational}. 

\begin{figure}[h]
\centering
\includegraphics[width=0.45\textwidth]{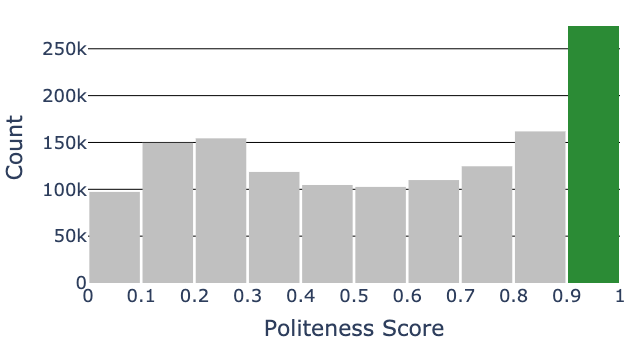}
\caption{\centering Distribution of Politeness Scores for the Enron Corpus}
\label{fig:distpol}
\end{figure}

\paragraph{Data Preparation}
The Enron corpus \cite{klimt2004introducing} consists of a large set of email conversations exchanged by the employees of the Enron corporation. Emails serve as a medium for exchange of requests, serving as an ideal application for politeness transfer.
We begin by pre-processing the raw Enron corpus following \citeauthor{shetty2004enron} \citeyearpar{shetty2004enron}. The first set of pre-processing\footnote{Pre-processing also involved steps for tokenization (done using spacy~\cite{spacy2}) and conversion to lower case.} steps and de-duplication yielded a corpus of roughly 2.5 million sentences. Further pruning\footnote{We prune the corpus by removing the sentences that 1) were less than 3 words long, 2) had more than 80\% numerical tokens, 3) contained email addresses, or 4) had repeated occurrences of spurious characters.} led to a cleaned corpus of over \datasetsize sentences. 
Finally, we use a politeness classifier \cite{niu-bansal-2018-polite} to assign politeness scores to these sentences and filter them into ten buckets based on the score (P\textsubscript{0}-P\textsubscript{9}; Fig. \ref{fig:distpol}). All the buckets are further divided into train, test, and dev splits (in a 80:10:10 ratio).

For our experiments, we assumed all the sentences with a politeness score of over 90\% by the classifier to be polite, also referred as the P\textsubscript{9} bucket (marked in green in Fig. \ref{fig:distpol}). 
We use the train-split of the P\textsubscript{9} bucket of over 270K polite sentences as the training data for the politeness transfer task.
Since the goal of the task is making action directives more polite, we manually curate a test set comprising of such sentences from test splits across the buckets.
We first train a classifier on the switchboard corpus \cite{Jurafsky-etal:1997} to get dialog state tags and filter sentences that have been labeled as either action-directive or quotation.\footnote{We used AWD-LSTM based classifier for classification of action-directive.}
Further, we use human annotators to manually select the test sentences. The annotators had a Fleiss's Kappa score ($\kappa$) of 0.77\footnote{The score was calculated for 3 annotators on a sample set of 50 sentences.} and curated a final test set of 800 sentences.

\begin{figure}[h]
\centering
\includegraphics[width=0.45\textwidth]{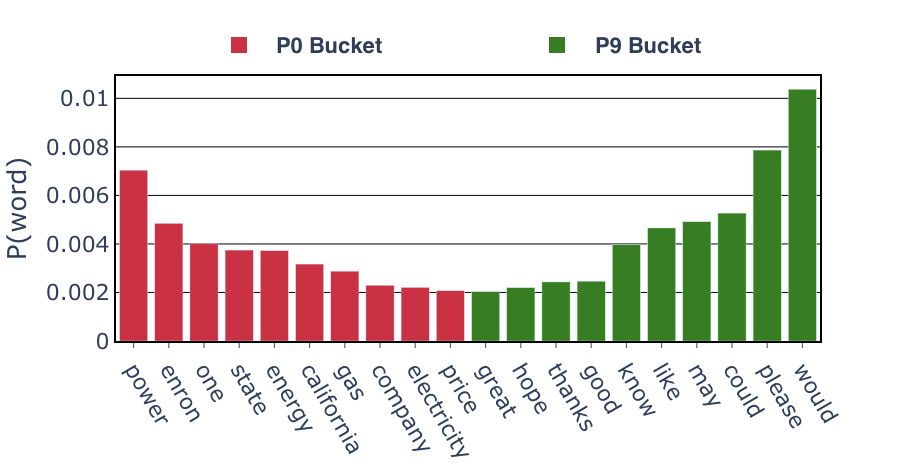}
\caption{\centering Probability of occurrence for 10 of the most common 30 words in the P\textsubscript{0} and P\textsubscript{9} data buckets}
\label{fig:top10_data}
\end{figure}

In Fig.~\ref{fig:top10_data}, we examine the two extreme buckets with politeness scores of $<10\%$ (P\textsubscript{0} bucket) and $>90\%$ (P\textsubscript{9} bucket) from our corpus by plotting 10 of the top 30 words occurring in each bucket.
We clearly notice that words in the P\textsubscript{9} bucket are closely linked to polite style, while words in the P\textsubscript{0} bucket are mostly content words. This substantiates our claim that the task of politeness transfer is fundamentally different from other attribute transfer tasks like sentiment where both the polarities are clearly defined.


\subsection{Other Tasks}
The \textbf{Captions} dataset \cite{gan2017stylenet} has image captions labeled as being factual, romantic or humorous. We use this dataset to perform transfer between these styles.
This task parallels the task of politeness transfer because much like in the case of politeness transfer, the captions task also involves going from a style neutral (factual) to a style rich (humorous or romantic) parlance.

For sentiment transfer, we use the \textbf{Yelp} restaurant review dataset \cite{shen2017style} to train, and evaluate on a test set of 1000 sentences released by \citeauthor{li-etal-2018-delete} \citeyearpar{li-etal-2018-delete}. We also use the \textbf{Amazon} dataset of product reviews \cite{he2016ups}.
We use the Yelp review dataset labelled for the \textbf{Gender} of the author, released by \citeauthor{prabhumoye-etal-2018-style} \citeyearpar{prabhumoye-etal-2018-style} compiled from  \citeauthor{reddy2016obfuscating} \citeyearpar{reddy2016obfuscating}.
For the \textbf{Political} slant task \cite{prabhumoye-etal-2018-style}, we use dataset released by \citeauthor{voigt-etal-2018-rtgender} \citeyearpar{voigt-etal-2018-rtgender}.
\section{Methodology}

We are given non-parallel samples of sentences $\V{X_1}=\{\V{x}_1^{(1)} \dots \V{x}_{n}^{(1)}\}$ and $\V{X_2}=\{\V{x}_1^{(2)} \dots \V{x}_{m}^{(2)}\}$  from styles $\C{S}_{1}$ and $\C{S}_{2}$ respectively. 
The objective of the task is to efficiently generate samples $\V{\hat{X}_1} = \{\V{\hat{x}}_1^{(2)} \dots \V{\hat{x}}_{n}^{(2)}\}$ in the target style $\C{S}_2$, conditioned on samples in $\V{X_1}$.  For a style $\C{S}_v$ where $v \in \{1,2\}$, we begin by learning a set of phrases ($\Gamma_{v}$) which characterize the style $\C{S}_v$. The presence of phrases from $\Gamma_v$ in a sentence $\V{x}_i$ would associate the sentence with the style $\C{S}_v$. For example, phrases like ``pretty good'' and ``worth every penny'' are characteristic of the ``positive'' style in the case of sentiment transfer task. 



\begin{figure*}[t!]
    \centering
    \includegraphics[width=\textwidth]{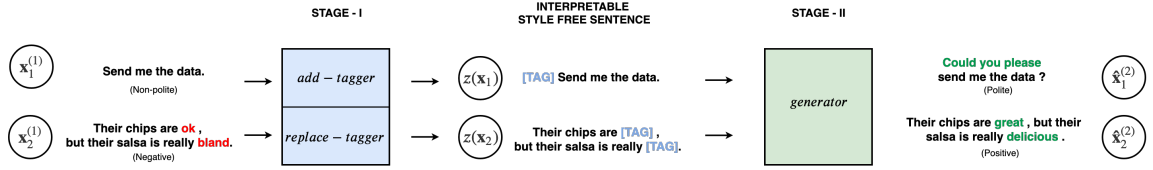}
    \caption{\centering Our proposed approach:  \textit{tag} and \textit{generate}. The tagger infers the interpretable style free sentence $z(\V{x}_i)$ for an input $\V{x}_i^{(1)}$ in source style $\C{S}_1$. The generator transforms  $\V{x}_i^{(1)}$ into  $\V{\hat{x}}_i^{(2)}$ which is in target style $\C{S}_2$.}
    \label{fig:model-pipeline}
\end{figure*}


We propose a two staged approach where we first infer a sentence $z(\V{x}_i)$ from $\V{x}_i^{(1)}$ using a model, the tagger.  The goal of the tagger is to ensure that the sentence $z(\V{x}_i)$ is agnostic to the original style ($\C{S}_{1}$) of the input sentence.
Conditioned on $z(\V{x}_i)$, we then generate the transferred sentence $\V{\hat{x}}^{(2)}_i$ in the target style $\C{S}_2$ using another model, the generator. The intermediate variable  $z(\V{x}_i)$ is also seen in other style-transfer methods. \citet{shen2017style, prabhumoye-etal-2018-style,yang2018unsupervised,pmlr-v70-hu17e} transform the input $\V{x}^{(v)}_i$ to a latent representation $z(\V{x}_i)$ which (ideally) encodes the content present in  $\V{x}^{(v)}_i$ while being agnostic to style $\C{S}_{v}$. In these cases $z(\V{x}_i)$ encodes the input sentence in a continuous latent space whereas for us $z(\V{x}_i)$ manifests in the surface form. 
The ability of our pipeline to generate observable intermediate outputs $z(\V{x}_i)$ makes it somewhat more interpretable than those other methods. 


We train two independent systems for the tagger \& generator which have complimentary objectives. 
The former identifies the style attribute markers $a(x^{(1)}_i)$ from source style  $\C{S}_1$ and either replaces them with a positional token called $\textsc{[tag]}$ or merely adds these positional tokens without removing any phrase from the input $x^{(1)}_i$. 
This particular capability of the model enables us to generate these tags in an input that is devoid of any attribute marker (i.e. $a(x^{(1)}_i)=\{\}$). 
This is one of the major differences from prior works which mainly focus on removing source style attributes and then replacing them with the target style attributes. It is especially critical for tasks like politeness transfer where the transfer takes place from a non-polite sentence.
This is because in such cases we may need to add new phrases to the sentence rather than simply replace existing ones. The generator is trained to generate sentences $\V{\hat{x}}_i^{(2)}$ in the target style by replacing these $\textsc{[tag]}$ tokens with stylistically relevant words inferred from target style $\C{S}_2$. 
Even though we have non-parallel corpora, both systems are trained in a supervised fashion as sequence-to-sequence models with their own distinct pairs of inputs \& outputs. 
To create parallel training data, we first estimate the style markers $\Gamma_v$ for a given style $\C{S}_v$ \& then use these to curate style free sentences with $[\textsc{tag}]$ tokens. 
Training data creation details are given in sections \Sref{sec:style-invariant}, \Sref{sec:targeted-generation}. 

Fig.~\ref{fig:model-pipeline} shows the overall pipeline of the proposed approach.
In the first example $\V{x}_1^{(1)}$, where there is no clear style attribute present, our model adds the $\textsc{[tag]}$ token in $z(\V{x}_1)$, indicating that a target style marker should be generated in this position. 
On the contrary, in the second example, the terms ``ok" and ``bland" are markers of negative sentiment and hence the tagger has replaced them with $\textsc{[tag]}$ tokens in $z(\V{x}_2)$.
We can also see that the inferred sentence in both the cases is free of the original and target styles. The structural bias induced by this two staged approach is helpful in realizing an interpretable style free tagged sentence that explicitly encodes the content. In the following sections we discuss in detail the methodologies involved in (1) estimating the relevant attribute markers for a given style, (2) tagger, and (3) generator modules of our approach.


\subsection{Estimating Style Phrases}
\label{subsec:est-tags}
Drawing from~\citet{li-etal-2018-delete}, we propose a simple approach based on n-gram tf-idfs to estimate the set $\Gamma_{v}$, which represents the style markers for style $v$. 
For a given corpus pair $\V{X}_{1}, \V{X}_2$ in styles $\C{S}_1, \C{S}_2$ respectively we first compute a probability distribution $p_{1}^{2}(w)$  over the n-grams $w$ present in both the corpora (Eq. \ref{eqn:tf-idf}). Intuitively, $p_{1}^{2}(w)$ is proportional to the probability of sampling an n-gram present in both $\V{X}_{1}, \V{X}_2$ but having a much higher tf-idf value in  $\V{X}_{2}$ relative to $\V{X}_1$. This is how we define the impactful style markers for style $\C{S}_2$. 

\begin{eqnarray}
    \label{eqn:tf-idf-1}
    & \eta_{1}^{2}(w) = \frac{\frac{1}{m}\sum\limits_{i=1}^{m} \textrm{tf-idf}(w, \V{x}_i^{(2)})}{\frac{1}{n}\sum\limits_{j=1}^{n} \textrm{tf-idf}(w, \V{x}_j^{(1)})} 
\end{eqnarray}
\begin{eqnarray}
    \label{eqn:tf-idf}
    & p_{1}^{2}(w) = \frac{\eta_{1}^{2}(w)^{\gamma}}{\sum\limits_{w'}\eta_{1}^{2}(w')^{\gamma}} 
\end{eqnarray}

where, $\eta_{1}^{2}(w)$ is the ratio of the mean tf-idfs for a given n-gram $w$ present in both $\V{X}_{1}, \V{X}_2$ with $|\V{X}_{1}|=n$ and $|\V{X}_{2}|=m$.
Words with higher values for $\eta_{1}^{2}(w)$ have a  higher mean tf-idf in $\V{X}_{2}$ vs $\V{X}_{1}$, and thus are more characteristic of $\C{S}_2$.
We further smooth and normalize $\eta_{1}^{2}(w)$ to get $p_{1}^{2}(w)$.
Finally, we estimate $\Gamma_{2}$ by

$$\Gamma_{2} = \{w : p_{1}^{2}(w) \geq k \}$$

In other words, $\Gamma_{2}$ consists of the set of phrases in $\V{X}_{2}$ above a given style impact $k$. $\Gamma_{1}$ is computed similarly where we use $p_{2}^{1}(w), \eta_{2}^{1}(w)$.

\subsection{Style Invariant Tagged Sentence} 
\label{sec:style-invariant}

The tagger model (with parameters $\theta_t$) takes as input the sentences in $\V{X}_1$ and outputs $\{z(\V{x}_i): \V{x}_i^{(1)} \in \V{X}_1\}$. Depending on the style transfer task, the tagger is trained to either (1) identify and replace style attributes $a(\V{x}_i^{(1)})$ with the token tag $\textsc{[tag]}$ (replace-tagger) or (2) add the $\textsc{[tag]}$ token at specific locations in $\V{x}_i^{(1)}$ (add-tagger). In both the cases, the $\textsc{[tag]}$ tokens indicate positions where the generator can insert phrases from the target style $\C{S}_2$. Finally, we  use the distribution $p_{1}^{2}(w)$/$p_{2}^{1}(w)$  over $\Gamma_2$/$\Gamma_1$  (\Sref{subsec:est-tags})  to draw samples of attribute-markers that would be replaced with the $\textsc{[tag]}$ token during the creation of training data. 


The first variant, replace-tagger, is suited for a task like sentiment transfer where almost every sentence has some attribute markers $a(\V{x}_i^{(1)})$ present in it. In this case the training data comprises of pairs where the input is $\V{X}_1$ and the output is $\{z(\V{x}_i):\V{x}_i^{(1)} \in \V{X}_1\}$. The loss objective for replace-tagger is given by $\C{L}_r(\theta_{t})$ in Eq. \ref{eqn:tagger-1}. 

\begin{eqnarray}
    \label{eqn:tagger-1}
    \C{L}_r(\theta_{t}) = -\sum_{i=1}^{|\V{X}_1|}\log P_{\theta_t}(z(\V{x}_i)|\V{x}_i^{(1)}; \theta_{t}) 
\end{eqnarray}

The second variant, add-tagger, is designed for cases where the transfer needs to happen from style neutral sentences to the target style. That is, $\V{X}_1$ consists of style neutral sentences whereas $\V{X}_2$ consists of sentences in the target style.  Examples of such a task include the tasks of politeness transfer (introduced in this paper) and caption style transfer (used by \citet{li-etal-2018-delete}). In such cases, since the source sentences have no attribute markers to remove, the tagger learns to add $\textsc{[tag]}$ tokens at specific locations suitable for emanating style words in the target style. 


\begin{figure}[h!]
    \centering
    \includegraphics[width=0.4\textwidth]{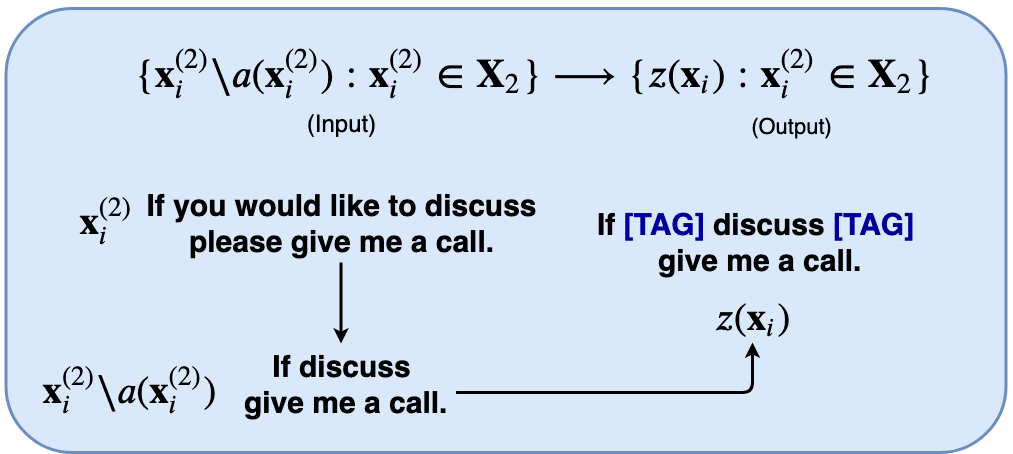}
    \caption{\centering Creation of training data for add-tagger.}
    \label{fig:add-tagger-training}
\end{figure}

The training data (Fig. \ref{fig:add-tagger-training}) for the add-tagger is given by pairs where the input is $\{ \V{x}_i^{(2)} \backslash a(\V{x}_i^{(2)}) : \V{x}_i^{(2)} \in \V{X}_2 \}$ and the output is $\{z(\V{x}_i):\V{x}_i^{(2)} \in \V{X}_2\}$. Essentially, for the input we take samples $\V{x}_i^{(2)}$ in the target style $\C{S}_2$ and explicitly remove style phrases $a(\V{x}_i^{(2)})$ from it. 
For the output we replace the same phrases $a(\V{x}_i^{(2)})$ with $\textsc{[tag]}$ tokens. As indicated in Fig. \ref{fig:add-tagger-training}, we remove the style phrases ``you would like to" and ``please" and replace them with $\textsc{[tag]}$ in the output. 
Note that we only use samples from $\V{X}_2$ for training the add-tagger; samples from the style neutral $\V{X}_1$ are not involved in the training process at all. For example, in the case of politeness transfer, we only use the sentences labeled as ``polite'' for training. In effect, by training in this fashion, the tagger learns to add $\textsc{[tag]}$ tokens at appropriate locations in a style neutral sentence. The loss objective ($\C{L}_{a}$) given by Eq. \ref{eqn:tagger-2} is crucial for tasks like politeness transfer where one of the styles is poorly defined.

\begin{eqnarray}
    \label{eqn:tagger-2}
    \C{L}_{a}(\theta_{t}) = -\sum_{i=1}^{|\V{X}_1|}\log P_{\theta_t}(z(\V{x}_i)|\V{x}_i^{(2)} \backslash a(\V{x}_i^{(2)}); \theta_{t})
\end{eqnarray}


\subsection{Style Targeted Generation}
\label{sec:targeted-generation}

The training for the generator model is complimentary to that of the tagger, in the sense that the  generator takes as input the tagged output $z(\V{x}_i)$ inferred from the source style and modifies the $\textsc{[tag]}$ tokens to generate the desired sentence $\V{\hat{x}}_i^{(v)}$ in the target style $\C{S}_v$.  

\begin{eqnarray}
    \label{eqn:generator-1}
    \C{L}(\theta_{g}) = -\sum_{i=1}^{|\V{X}_v|}\log P_{\theta_g}(\V{x}_i^{(v)}|z(\V{x}_i); \theta_{g}) 
\end{eqnarray}


The training data for transfer into style $\C{S}_v$ comprises of pairs where the input is given by $\{z(\V{x}_i) : {x}_i^{(v)} \in \V{X}_v \,,\, v \in \{1,2\}\}$ and the output is $\V{X}_v$, i.e. it is trained to transform a style agnostic representation into a style targeted sentence. Since the generator has no notion of the original style and it is only concerned with the style agnostic representation $z(\V{x}_i)$, it is convenient to disentangle the training for tagger \& generator. 

Finally, we note that the location at which the tags are generated has a significant impact on the distribution over style attributes (in $\Gamma_2$) that are used to fill the $\textsc{[tag]}$ token at a particular position.
Hence, instead of using a single $\textsc{[tag]}$ token, we use a set of positional tokens $\textsc{[tag]}_{t}$ where $t \in \{0, 1, \dots\ T \}$ for a sentence of length $T$.
By training both tagger and generator with these positional $\textsc{[tag]}_{t}$  tokens we enable them to easily realize different distributions of style attributes for different positions in a sentence. 
For example, in the case of politeness transfer, the tags added at the beginning ($t=0$) will almost always be used to generate a token like ``Would it be possible ..." whereas for a higher $t$, $\textsc{[tag]}_{t}$  may be replaced with a token like ``thanks" or ``sorry." 



\begin{table*}[t]
\centering
\setlength{\tabcolsep}{0.25em} 
\small{
\begin{tabular}{c@{\hskip 0.3in} r r r r@{\hskip 0.3in} r r r r @{\hskip 0.3in} r r r r} 
\toprule
     & \multicolumn{4}{c}{\textbf{Politeness}}{\hskip 0.3in} & \multicolumn{4}{c}{\textbf{Gender}}{\hskip 0.3in} & \multicolumn{4}{c}{\textbf{Political}}  \\ 
\midrule
     & Acc   & BL-s  & MET & ROU                               & Acc   & BL-s  & MET & ROU                            & ACC   & BL-s  & MET & ROU                               \\ 
\midrule
CAE  & \textbf{99.62} & 6.94  & 10.73  & 25.71                               & 65.21 & 9.25  & 14.72  & 42.42                            & 77.71 & 3.17  & 7.79   & 27.17                               \\ 
BST  & 60.75 & 2.55  & 9.19   & 18.99                               & 54.4  & 20.73 & 22.57  & 55.55                            & \textbf{88.49} & 10.71 & 16.26  & 41.02                               \\
DRG  & 90.25 & 11.83 & 18.07  & 41.09                               & 36.29 & 22.9  & 22.84  & 53.30                            & 69.79     & 25.69     & 21.6      & 51.8                                   \\ 
OURS & 89.50 & \textbf{70.44} &  \textbf{36.26}  &  \textbf{70.99} & 
 \textbf{82.21} &  \textbf{52.76} &  \textbf{37.42}  &  \textbf{74.59 }                      
 & 87.74 & \textbf{68.44} & \textbf{45.44}  & \textbf{77.51}                               \\ 
\midrule
\end{tabular}
}
\vspace{-0.8em}
\caption{\centering Results on the Politeness, Gender and Political datasets.}
\vspace{-0.5em}
\label{tab:polite}
\end{table*}


\section{Experiments and Results}

\label{sec:expt_results}

\paragraph{Baselines}
We compare our systems against three previous methods. \textsc{drg} \cite{li-etal-2018-delete}, Style Transfer Through Back-translation (\textsc{bst}) \cite{prabhumoye-etal-2018-style}, and Style transfer from non-parallel text by cross alignment \cite{shen2017style} (\textsc{cae}). 
For \textsc{drg}, we only compare against the best reported method, delete-retrieve-generate.
For all the models, we follow the experimental setups described in their respective papers.

\paragraph{Implementation Details}

We use 4-layered transformers \cite{vaswani2017attention} to train both tagger and generator modules. Each transformer has $4$ attention heads with a $512$ dimensional embedding layer and hidden state size. Dropout \cite{srivastava2014dropout} with p-value $0.3$ is added for each layer in the transformer. For the politeness dataset the generator module is trained with data augmentation techniques like random word shuffle, word drops/replacements as proposed by~\cite{im2017denoising}. We empirically observed that these techniques provide an improvement in the fluency and diversity of the generations. Both modules were also trained with the BPE tokenization \cite{sennrich2015neural} using a vocabulary of size $16000$ for all the datasets except for Captions, which was trained using $4000$ BPE tokens. 
The value of the smoothing parameter $\gamma$ in Eq. \ref{eqn:tf-idf} is set to $0.75$. For all datasets except Yelp we use phrases with  $p_{1}^{2}(w)$ $\geq k=0.9$ to construct $\Gamma_2$, $\Gamma_1$ (\Sref{subsec:est-tags}). For Yelp $k$ is set to $0.97$. During inference we use beam search (beam size=5) to decode tagged sentences and targeted generations for tagger \& generator respectively. For the tagger, we re-rank the final  beam search outputs based on the number of $\textsc{[tag]}$ tokens in the output sequence (favoring more $\textsc{[tag]}$ tokens). 

\begin{table*}[t]

\centering
\setlength{\tabcolsep}{0.25em} 

\small{
\begin{tabular}{c@{\hskip 0.3in} r r r r r@{\hskip 0.3in} r r r r r @{\hskip 0.3in} r r r r r } 
\toprule
     & \multicolumn{5}{c}{\textbf{Yelp }}{\hskip 0.3in} & \multicolumn{5}{c}{\textbf{Amazon }}{\hskip 0.3in} & \multicolumn{5}{c}{\textbf{Captions }}  \\ 
\midrule
     & Acc  & BL-s & BL-r & MET & ROU                           & Acc  & BL-s & BL-r & MET & ROU                          & Acc   & BL-s & BL-r & MET & ROU                          \\ 
\midrule
CAE  & 72.1 & 19.95  & 7.75   & 21.70  & 55.9                            & 78   & 2.64   & 1.68   & 9.52   & 29.16                          & 89.66 & 2.09   & 1.57   & 9.61   & 30.02                          \\ 
DRG  & \textbf{88.8} & 36.69  & 14.51  & 32.09  & 61.06                           & 52.2 & 57.07  & 29.85  & \textbf{50.16}  & 79.31                          & \textbf{95.65} & 31.79  & 11.78  & 32.45  & 64.32                          \\ 
OURS & 86.6 & \textbf{47.14}  & \textbf{19.76}  & \textbf{36.26}  & \textbf{70.99}                           & \textbf{66.4} & \textbf{68.74}  & \textbf{34.80}  & 45.3 & \textbf{83.45}                         & 93.17 & \textbf{51.01}  & \textbf{15.63}  & \textbf{43.67}  & \textbf{79.51}                         \\
\midrule
\end{tabular}
}
\vspace{-0.8em}
\caption{\centering Results on the Yelp, Amazon and Captions datasets.}
\vspace{-1.5em}
\label{tab:yelp}
\end{table*}

\paragraph{Automated Evaluation}

Following prior work \cite{li-etal-2018-delete, shen2017style}, we use automatic metrics for evaluation of the models along two major dimensions: (1) style transfer accuracy and (2) content preservation. To capture accuracy, we use a classifier trained on the nonparallel style corpora for the respective datasets (barring politeness). The architecture of the classifier is based on \textsc{awd-lstm} \cite{merity2017regularizing} and a softmax layer trained via cross-entropy loss. We use the implementation provided by fastai.\footnote{\url{https://docs.fast.ai/}}
For politeness, we use the classifier trained by \cite{niu-bansal-2018-polite}.\footnote{This is trained on the dataset given by \cite{danescu-niculescu-mizil-etal-2013-computational}.} The metric of transfer accuracy \textbf{(Acc)} is defined as the percentage of generated sentences classified to be in the target domain by the classifier. The standard metric for measuring content preservation is \textsc{bleu}-self \textbf{(BL-s)} \cite{papineni2002bleu} which is computed with respect to the original sentences. Additionally, we report the \textsc{bleu}-reference \textbf{(BL-r)} scores using the human reference sentences on the Yelp, Amazon and Captions datasets \cite{li-etal-2018-delete}.
We also report \textsc{rouge} \textbf{(ROU)} \cite{lin2004rouge} and \textsc{meteor} \textbf{(\textsc{met})} \cite{denkowski2011meteor} scores. In particular, \textsc{meteor} also uses synonyms and stemmed forms of the words in candidate and reference sentences, and thus may be better at quantifying semantic similarities.

Table \ref{tab:polite} shows that our model achieves significantly higher scores on \textsc{bleu}, \textsc{rouge} and \textsc{meteor} as compared to the baselines \textsc{drg}, \textsc{cae} and \textsc{bst} on the Politeness, Gender and Political datasets. The \textsc{bleu} score on the Politeness task is greater by $58.61$ points with respect to \textsc{drg}. 
In general, \textsc{cae} and \textsc{bst} achieve high classifier accuracies but they fail to retain the original content. 
The classifier accuracy on the generations of our model are comparable (within $1\%$) with that of \textsc{drg} for the Politeness dataset.

In Table \ref{tab:yelp}, we compare our model against \textsc{cae} and \textsc{drg} on the Yelp, Amazon, and Captions datasets. For each of the datasets our test set comprises 500 samples (with human references) curated by \citet{li-etal-2018-delete}. We observe an increase in the \textsc{bleu}-reference scores by $5.25$, $4.95$ and $3.64$ on the Yelp, Amazon, and Captions test sets respectively.
Additionally, we improve the transfer accuracy for Amazon by $14.2\%$ while achieving accuracies similar to \textsc{drg} on Yelp and Captions. 
As noted by \citet{li-etal-2018-delete}, one of the unique aspects of the Amazon dataset is the absence of similar content in both the sentiment polarities. Hence, the performance of their model is worse in this case. Since we don't make any such assumptions, we perform significantly better on this dataset.

While popular, the metrics of transfer accuracy and \textsc{bleu} have significant shortcomings making them susceptible to simple adversaries. \textsc{bleu} relies heavily on n-gram overlap and classifiers can be fooled by certain polarizing keywords. We test this hypothesis on the sentiment transfer task by a \textit{Naive Baseline}. This baseline adds \textit{``but overall it sucked"} at the end of the sentence to transfer it to negative sentiment. Similarly, it appends \textit{``but overall it was perfect"} for transfer into a positive sentiment. This baseline achieves an average accuracy score of $91.3\%$ and a \textsc{bleu} score of $61.44$ on the Yelp dataset. Despite high evaluation scores, it does not reflect a high rate of success on the task. In summary, evaluation via automatic metrics might not truly correlate with task success.

\paragraph{Changing Content Words}
Given that our model is explicitly trained to generate new content only in place of the TAG token, it is expected that a well-trained system will retain most of the non-tagged (content) words. Clearly, replacing content words is not desired since it may drastically change the meaning. In order to quantify this, we calculate the fraction of non-tagged words being changed across the datasets. We found that the non-tagged words were changed for only 6.9\% of the sentences. In some of these cases, we noticed that changing non-tagged words helped in producing outputs that were more natural and fluent.

\begin{table}[t]

\centering
\setlength{\tabcolsep}{0.25em} 

\small{
\begin{tabular}{l @{\hskip 0.1in} r r @{\hskip 0.1in} r r @{\hskip 0.1in} r r } 
\toprule
     & \multicolumn{2}{c}{\textbf{Con}}{\hskip 0.1in} & \multicolumn{2}{c}{\textbf{ Att}}{\hskip 0.1in} & \multicolumn{2}{c}{\textbf{Gra}}  \\ 
\midrule
     & DRG & Ours   & DRG & Ours &   DRG & Ours   \\ 
\midrule

Politeness & 2.9 & \textbf{3.6} & 3.2 & \textbf{3.6} & 2.0 & \textbf{3.7} \\
Gender & 3.0 & \textbf{3.5} & - & - & 2.2 & \textbf{2.5} \\
Political & 2.9 & \textbf{3.2} & - & - & 2.5 & \textbf{2.7} \\
Yelp & 3.0 & \textbf{3.7} & 3 & \textbf{3.9} & 2.7 & \textbf{3.3} \\

\midrule
\end{tabular}
}
\vspace{-0.8em}
\caption{\centering Human evaluation on Politeness, Gender, Political and Yelp datasets. }
\vspace{-0.3em}
\label{tab:human-eval}
\end{table}

\begin{table*}[!ht]
\centering
\small{
\begin{tabular}{p{1.5in}@{\hskip 0.2in} p{1.5in}@{\hskip 0.2in}p{1.6in}@{\hskip 0.2in}p{0.75in}} 
\toprule
\multicolumn{1}{c}{\textbf{Input}}  & \multicolumn{1}{c}{\textbf{\textsc{drg} Output}} & \multicolumn{1}{c}{\textbf{Our Model Output}} & \multicolumn{1}{c}{\textbf{\thead{Strategy}}} \\
\midrule
what happened to my personal station? & what happened to my mother to my co??? & \textcolor{green!65!blue}{could you please let me know} what happened to my personal station? & \thead{Counterfactual \\ Modal} \\
\addlinespace
yes, go ahead and remove it. & yes, please go to the link below and delete it. & yes, \textcolor{green!65!blue}{we can} go ahead and remove it. & 1st Person Plural \\
\addlinespace
not yet-i'll try this wkend. & not yet to say-i think this will be a \textless{}unk\textgreater{} long. & \textcolor{green!65!blue}{sorry} not yet-i'll try to make sure this wk & Apologizing \\
\addlinespace
please check on metromedia energy, & thanks again on the energy industry, & please check on metromedia energy, \textcolor{green!65!blue}{thanks} & Mitigating please start \\
\midrule
\end{tabular}}
\caption{Qualitative Examples comparing the outputs from \textsc{drg} and Our model for the Politeness Transfer Task}

\label{tab:qual_res}
\end{table*}

\paragraph{Human Evaluation}
\label{subsec:human-eval}
Following \citet{li-etal-2018-delete}, we select 10 unbiased human judges to rate the output of our model and \textsc{drg} on three aspects: (1) content preservation \textbf{(Con)} (2) grammaticality of the generated content \textbf{(Gra)} (3) target attribute match of the generations \textbf{(Att)}. For each of these metrics, the reviewers give a score between 1-5 to each of the outputs, where 1 reflects a poor performance on the task and 5 means a perfect output. Since the judgement of signals that indicate gender and political inclination are prone to personal biases, we don't annotate these tasks for target attribute match metric.
Instead we rely on the classifier scores for the transfer. We've used the same instructions from \citet{li-etal-2018-delete} for our human study. Overall, we evaluate both systems on a total of 200 samples for Politeness and 100 samples each for Yelp, Gender and Political.

Table \ref{tab:human-eval} shows the results of human evaluations. We observe a significant improvement in content preservation scores across various datasets (specifically in Politeness domain) highlighting the ability of our model to retain content better than \textsc{drg}. Alongside, we also observe consistent improvements of our model on target attribute matching and grammatical correctness. 

\paragraph{Qualitative Analysis}

We compare the results of our model with the \textsc{drg} model qualitatively as shown in Table \ref{tab:qual_res}. Our analysis is based on the linguistic strategies for politeness as described in \cite{danescu-niculescu-mizil-etal-2013-computational}. The first sentence presents a simple example of the \textit{counterfactual modal} strategy inducing \textit{``Could you please"} to make the sentence polite. The second sentence highlights another subtle concept of politeness of \textit{1st Person Plural} where adding \textit{``we"} helps being indirect and creates the sense that the
burden of the request is shared between speaker
and addressee. The third sentence highlights the ability of the model to add \textit{Apologizing} words like \textit{``Sorry"} which helps in deflecting the social threat of the request by attuning to the imposition. According to the \textit{Please Start} strategy, it is more direct and insincere to start a sentence with \textit{``Please"}. The fourth sentence projects the case where our model uses \textit{``thanks"} at the end to express gratitude and in turn, makes the sentence more polite.
Our model follows the strategies prescribed in \cite{danescu-niculescu-mizil-etal-2013-computational} while generating polite sentences.\footnote{We provide additional qualitative examples  for other tasks in the supplementary material.}

\paragraph{Ablations}
We provide a comparison of the two variants of the tagger, namely the replace-tagger and add-tagger on two datasets.
We also train and compare them with a \textit{combined} variant.\footnote{Training of combined variant is done by training the tagger model on the concatenation of training data for add-tagger and replace-tagger.}
We train these tagger variants on the Yelp and Captions datasets and present the results in Table \ref{tab:ablations}. 
We observe that for Captions, where we transfer a factual (neutral) to romantic/humorous sentence, the add-tagger provides the best accuracy with a relatively negligible drop in \textsc{bleu} scores. On the contrary, for Yelp, where both polarities are clearly defined, the replace-tagger gives the best performance. 
Interestingly, the accuracy of the add-tagger is $\approx 50\%$ in the case of Yelp, since adding negative words to a positive sentence or vice-versa neutralizes the classifier scores. Thus, we can use the add-tagger variant for transfer from a polarized class to a neutral class as well.

To check if the combined tagger is learning to perform the operation that is more suitable for a dataset, we calculate the fraction of times the combined tagger performs add/replace operations on the Yelp and Captions datasets. We find that for Yelp (a polar dataset) the combined tagger performs 20\% more replace operations (as compared to add operations). In contrast, on the \textsc{captions} dataset, it performs 50\% more add operations. While the combined tagger learns to use the optimal tagging operation to some extent, a deeper understanding of this phenomenon is an interesting future topic for research.
We conclude that the choice of the tagger variant is dependent on the characterstics of the underlying transfer task.

\begin{table}[!h]
\centering
\footnotesize{
\setlength{\tabcolsep}{2.5pt}
\begin{tabular}{l@{\hskip 0.3in}rr@{\hskip 0.3in}rr} 
\toprule
 & \multicolumn{2}{c}{\textbf{Yelp}}{\hskip 0.3in} & \multicolumn{2}{c}{\textbf{Captions}}  \\ 
\midrule
                & Acc  & BL-r      & Acc & BL-r          \\ 
\midrule
Add-Tagger      & 53.2 & 20.66          &  \textbf{93.17}  &   15.63    \\
Replace-Tagger  &  \textbf{86.6}    &   19.76        & 84.5  &   15.04   \\
Combined & 72.5  &   \textbf{22.46}     & 82.17    &   \textbf{18.5}1   \\
\midrule
\end{tabular}
}
\caption{\centering Comparison of different \textit{tagger} variants for Yelp and Captions datasets}
\label{tab:ablations}
\end{table}

\section{Conclusion}
We introduce the task of politeness transfer for which we provide a dataset comprised of sentences curated from email exchanges present in the Enron corpus. 
We extend prior works \cite{li-etal-2018-delete, sudhakar-etal-2019-transforming} on attribute transfer by introducing a simple pipeline -- $tag$ \& $generate$ which is an interpretable two-staged approach for content preserving style transfer. 
We believe our approach is the first to be robust in cases when the source is style neutral, like the ``non-polite" class in the case of politeness transfer. 
Automatic and human evaluation shows that our approach outperforms other state-of-the-art models on content preservation metrics while retaining (or in some cases improving) the transfer accuracies. 





\section*{Acknowledgments}
This material is based on research sponsored in part by the Air Force Research Laboratory under agreement number FA8750-19-2-0200. 
The U.S. Government is authorized to reproduce and distribute reprints for Governmental purposes notwithstanding any copyright notation thereon. 
The views and conclusions contained herein are those of the authors and should not be interpreted as necessarily representing the official policies or endorsements, either expressed or implied, of the Air Force Research Laboratory or the U.S. Government. 
This work was also supported in part by ONR Grant N000141812861, NSF IIS1763562, and Apple.
We would also like to acknowledge NVIDIA's GPU support.
We would like to thank Antonis Anastasopoulos, Ritam Dutt, Sopan Khosla, and, Xinyi Wang for the helpful discussions.

\bibliography{acl2020}
\bibliographystyle{acl_natbib}

\appendix

\begin{table*}
\centering
\small{
\begin{tabular}{p{1.86in}@{\hskip 0.3in} p{1.86in}@{\hskip 0.3in} p{1.86in}} 
\toprule
\multicolumn{1}{c}{\textbf{Non-polite Input}}  & \multicolumn{1}{c}{\textbf{DRG}} & \multicolumn{1}{c}{\textbf{Our Model}} \\ 
\midrule
jon - - please use this resignation letter in lieu of the one sent on friday . & - i think this would be a good idea if you could not be a statement that harry 's signed in one of the schedule . & jon - \textcolor{green!65!blue}{sorry} - please use this resignation letter in lieu of the one event sent on \\
\addlinespace
if you have a few minutes today, give me a call & i'll call today to discuss this. & if you have a few minutes today, \textcolor{green!65!blue}{please} give me a call at \\
\addlinespace
anyway you can let me know.& anyway, i'm sure i'm sure. & anyway \textcolor{green!65!blue}{please} let me know as soon as possible \\ 
\addlinespace
yes, go ahead and remove it. & yes, please go to the link below and delete it. & yes, \textcolor{green!65!blue}{we can} go ahead and remove it.\\ 
\addlinespace
can you explain a bit more about how those two coexist ? also ..... & i can explain how the two more than \textless{}unk\textgreater{} i can help with mike ? & can you explain a bit more about how those two coexist ? also \textcolor{green!65!blue}{thanks} \\
\addlinespace
go ahead and sign it - i did . &  go away so we can get it approved .	& \textcolor{green!65!blue}{we could} go ahead and sign it - i did look at \\
\midrule
\end{tabular}}
\caption{\centering Additional Qualitative Examples of outputs from our Model and DRG for the Politeness Transfer Task}
\end{table*}

\begin{table*}
\centering
\small{
\begin{tabular}{p{0.75in}|p{1.65in}|p{1.65in}|p{1.65in}} 
\hline
\textbf{Task} & \textbf{Non-polite Input}  & \textbf{DRG}                                                          & \textbf{Our Model}                                          \\ 
\hline \hline
Fem $\rightarrow$ Male & my husband ordered the brisket . & my wife had the best steak . & my \textcolor{green!65!blue}{wife} ordered the brisket . \\
\hline
Fem $\rightarrow$ Male & i ' m a fair person . & i ' m a good job of the \textless{}unk\textgreater{} . & i ' m a \textcolor{green!65!blue}{big guy} . \\ 
\hline
Male $\rightarrow$ Fem & my girlfriend and i recently stayed at this sheraton . & i recently went with the club . & my \textcolor{green!65!blue}{husband} and i recently stayed at this office . \\ 
\hline
Male $\rightarrow$ Fem & however , once inside the place was empty . & however , when the restaurant was happy hour for dinner . & however , once inside the place was \textcolor{green!65!blue}{super cute} . \\ 
\hline
Pos $\rightarrow$ Neg & good drinks , and good company . & horrible company . & \textcolor{green!65!blue}{terrible} drinks , \textcolor{green!65!blue}{terrible} company.\\ 
\hline
Pos $\rightarrow$ Neg & i will be going back and enjoying this great place ! & i will be going back and enjoying this great ! & i will \textcolor{green!65!blue}{not} be going back and enjoying this \textcolor{green!65!blue}{garbage} ! \\
\hline
Neg $\rightarrow$ Pos & this is the reason i will never go back . & this is the reason i will never go back . & so happy i will \textcolor{green!65!blue}{definitely be back} . \\
\hline
Neg $\rightarrow$ Pos & salsa is not hot or good . & salsa is not hot or good . & salsa is \textcolor{green!65!blue}{always} hot and \textcolor{green!65!blue}{fresh} .\\
\hline
Dem $\rightarrow$ Rep & i am confident of trumps slaughter . & i am mia love & i am confident of trumps \textcolor{green!65!blue}{administration} . \\
\hline
Dem $\rightarrow$ Rep & we will resist trump & we will impeach obama & we will be \textcolor{green!65!blue}{praying} for trump \\
\hline
Rep $\rightarrow$ Dem & video : black patriots demand impeachment of obama & video : black police show choose & video : black patriots demand to \textcolor{green!65!blue}{endorse} obama \\
\hline
Rep $\rightarrow$ Dem & mr. trump is good ... but mr. marco rubio is great ! ! & thank you mr. good ... but mr. kaine is great senator ! ! & mr. \textcolor{green!65!blue}{schumer} is good ... but mr. \textcolor{green!65!blue}{pallone} is great ! ! \\
\hline
Fact $\rightarrow$ Rom & a woman is sitting near a flower bed overlooking a tunnel . & a woman is sitting near a flower overlooking a tunnel, determined to & a woman is sitting near a brick rope , \textcolor{green!65!blue}{excited to meet her boyfriend} . \\
\hline
Fact $\rightarrow$ Rom & two dogs play with a tennis ball in the snow . & two dogs play with a tennis ball in the snow . & two dogs play with a tennis ball in the snow \textcolor{green!65!blue}{celebrating their friendship} . \\
\hline
Fact $\rightarrow$ Hum & three kids play on a wall with a green ball . & three kids on a bar on a field of a date . & three kids play on a wall with a green ball \textcolor{green!65!blue}{fighting for supremacy} . \\
\hline
Fact $\rightarrow$ Hum & a black dog plays around in water . & a black dog plays in the water . & a black dog plays around in water \textcolor{green!65!blue}{looking for fish} . \\
\hline
\end{tabular}}
\caption{ \centering Additional Qualitative Examples of our Model and DRG for other Transfer Tasks}
\end{table*}

\end{document}